\pdfoutput=1

\documentclass[11pt]{article}

\usepackage[preprint]{acl}
\usepackage[title]{appendix}
\usepackage{times}
\usepackage{latexsym}
\usepackage{subcaption}
\usepackage{enumitem}

\usepackage[T1]{fontenc}

\usepackage[utf8]{inputenc}

\usepackage{microtype}

\usepackage{inconsolata}

\usepackage{graphicx}
\usepackage{amsmath}    
\usepackage{array}      
\usepackage{booktabs}   
\usepackage{multirow}   
\usepackage{amssymb}
\usepackage{pifont} 
\usepackage{xcolor,colortbl} 
\usepackage{float}
\usepackage{tabularx}
\usepackage[ruled,vlined]{algorithm2e}

\title{SQLCritic: Correcting Text-to-SQL Generation via Clause-wise Critic}

\author{
    Jikai Chen, Leilei Gan, Ziyu Zhao, Zechuan Wang \\ 
    Zhejiang University \\ 
    \texttt{\{jikaichen, leileigan, ziyuzhao.cs, 22451331\}@zju.edu.cn} 
    \AND 
    Dong Wang, Chenyi Zhuang \\ 
    Ant Group \\ 
    \texttt{\{yishan.wd, chenyi.zcy\}@antgroup.com}
}

\begin{document}
\maketitle
\begin{abstract}
Existing refinement methods in LLM-based Text-to-SQL systems exhibit limited effectiveness. 
They often introduce new errors during the self-correction process and fail to detect and correct semantic inaccuracies.
To address these gaps, we first introduce a clause-wise critique generation task along with a benchmark, SQLCriticBench, which performs fine-grained error localization including both syntax and semantic errors at the clause level.
Furthermore, we introduce a variant of DPO for training our SQLCritic model, where the $\beta$ coefficient is adaptively changed according to the clause-level inconsistencies between the preferred and dispreferred critiques.
We also propose an automatically training dataset curation pipeline which annotate clause-wise critique at scale in a cost-effective way.
Experiments demonstrate that the SQLCritic model significantly improves SQL accuracy on the BIRD and Spider datasets, and the results on SQLCriticBench further reveals its superior critique capabilities compared to existing models. 
\end{abstract}

\definecolor{task_blue}{RGB}{95,145,215}
\definecolor{knowledge_red}{RGB}{178,78,83}
\definecolor{general_green}{RGB}{179,214,163}

\section{Introduction}
In recent years, large language model (LLM)-based Text-to-SQL (Text2SQL) generation has seen significant advancements, substantially advancing the state of the art in translating natural language queries into SQL queries~\cite{hong2025nextgenerationdatabaseinterfacessurvey, zhu2024largelanguagemodelenhanced, qin2022surveytexttosqlparsingconcepts,dinsql,dailsql,macsql}.

\begin{figure}[t]
    \centering
    \includegraphics[width=0.95\linewidth]{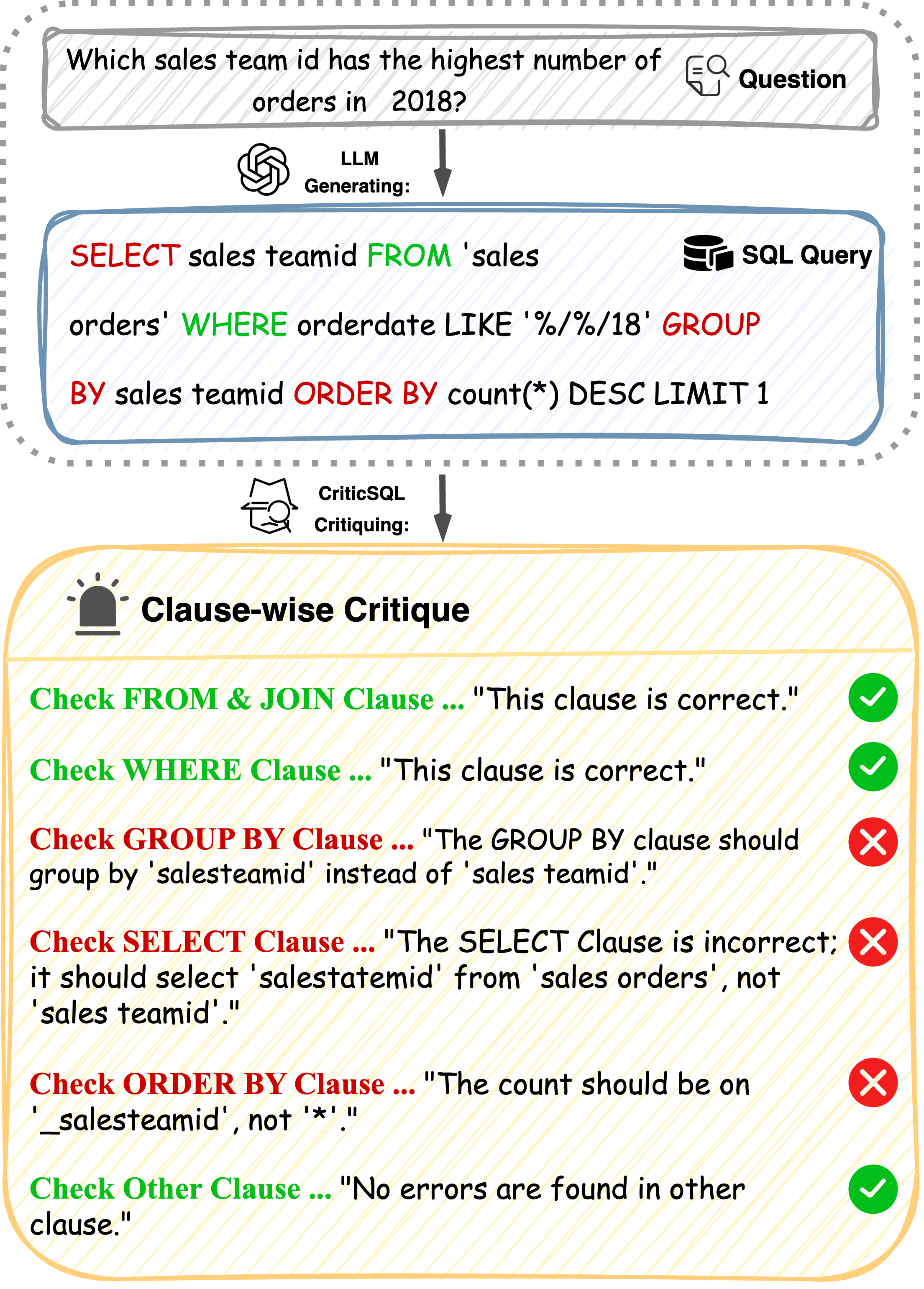}
    \caption{Clause-wise critique in the Text-to-SQL critique and correction task.}
    \label{fig:framework}
\end{figure}

The development of LLM-based Text2SQL generation can be divided into the following stages. 
Initially, most studies focus on designing Text2SQL-specific prompts for LLMs (e.g., DIN-SQL~\cite{dinsql} and DAIL-SQL~\cite{dailsql}).
For instance, they decompose the Text2SQL task into subtasks—schema linking, SQL generation, and SQL correction—and develop tailored prompts for each stage accordingly. 
Beyond directly prompting LLMs, subsequent studies explore multi-agent collaboration frameworks for Text2SQL generation~\cite{multi-agent}, in which different LLMs are assigned specific roles of the task and collaborate to produce high-quality SQL queries.
In addition to exploiting proprietary LLMs for Text2SQL generation, recent studies have begun developing Text2SQL-specific LLMs by continuing training open-source LLMs on high-quality text-to-sql corpora, achieving state-of-the-art results~\cite{li2024codes,starcoder,sun2023sql}.

In modern LLM-based Text2SQL generation methods, the critique and correction components—which identify and rectify generation errors—play a vital role in advancing practical and interactive applications.
However, existing methods that can generally be categorized into self-correction and execution feedback-based approaches remain limited in their effectiveness.
First, for self-correction~\cite{dinsql, dong2023c3, wang2022self}, they iteratively refine outputs using chain-of-thought reasoning or self-consistency, but  frequently modify previously correct results, thus introducing new errors~\cite{huang2023large, kamoi2024can}. 
On the other hand, execution feedback-based methods~\cite{macsql} effectively detect and correct syntactic errors to ensure the generated SQL can be executed successfully.
However, they fail to correct those SQLs with semantic errors which constitute the majority of errors in generated outputs~\cite{magic}, thereby limiting the overall effectiveness and practical applicability.

To address these challenges, we first introduce a clause-wise critique generation task along with a benchmark, \textbf{SQLCriticBench}, which performs fine-grained error localization including both \textbf{syntax} and \textbf{semantic errors} at the clause level, as shown in Figure~\ref{fig:framework}.
Compared to response-level critiques, clause-wise critiques better reflect the compositional structure of SQL query, enabling precise error localization and clearer interpretation. 
They also provide more actionable feedback for LLM-based correction and support more accurate automatic evaluation against human annotations.
Furthermore, we introduce a variant of Direct Preference Optimization (DPO;~\cite{dpo}) for training our SQLCritic model, where the $\beta$ coefficient is adaptively changed according to the clause-level inconsistencies between the preferred and dispreferred critiques.
This dynamic $\beta$ coefficient helps adjust the update steps when encountering different preference pairs.
Lastly, we propose an automatically training dataset curation pipeline which annotate clause-wise critique at scale in a cost-effective way.

To assess the effectiveness of SQLCritic, we apply it to critique and correct the predicted SQL queries generated by state-of-the-art Text2SQL systems, achieving absolute performance gains ranging from 2\% to 8\%. 
We further evaluate SQLCritic on the proposed SQLCriticBench dataset, where it demonstrates superior critique capabilities compared to leading closed-source models such as GPT-4~\cite{gpt4}.
These results highlight the potential of SQLCritic towards more realistic and interactive Text2SQL applications.

\section{Related Work}
\paragraph{Text-to-SQL Generation.} Early Text-to-SQL systems rely on rule-based or template-driven methods, which struggle with complex schemas \cite{li2014constructing,mahmud2015rule}. With advances in deep neural networks and pre-trained language models (PLMs), data-driven approaches markedly improve natural language-to-SQL mappings \cite{sutskever2014sequence,yin2020tabert}. Recent LLM-based methods like DAIL-SQL \cite{dailsql} and DIN-SQL \cite{dinsql} utilize techniques like chain-of-thought reasoning \cite{wei2022chain} and least-to-most prompting \cite{zhou2022least}, achieving notable accuracy. However, reliance on proprietary models raises concerns about privacy and cost, while open-source options like CodeS \cite{li2024codes} and SQL-PaLM \cite{sun2023sql} show potential but struggle with semantic accuracy for complex SQLs. Critiquing methods in LLM-based Text-to-SQL generation typically rely on execution feedback~\cite{macsql} or self-feedback~\cite{magic} to identify and correct syntactic and semantic errors. However, these approaches often yield suboptimal results. To address this limitation, we propose a clause-wise critique generation task supported by a well-trained critic model.

\paragraph{LLMs Critique.} LLM-based critique methods aim to refine model outputs~\cite{deepcritic, llmcritique, mcaleese2024llmcriticshelpcatch, pan2024automatically}. These approaches can be categorized as either \textit{intrinsic}, leveraging techniques such as chain-of-thought reasoning~\cite{cot, t2i-r1}, self-consistency~\cite{self-consistency}, and reflection~\cite{shinn2024reflexion,tyen2023llms}; or \textit{extrinsic}, which utilize external signals such as execution results~\cite{huang2023large, xiao2025detectingmitigatinghallucinationlarge}. 

In this work, we propose a clause-wise critique generation task supported by a well-trained critic model.

\begin{figure*}[t]
    \centering
    \includegraphics[width=0.9\linewidth]{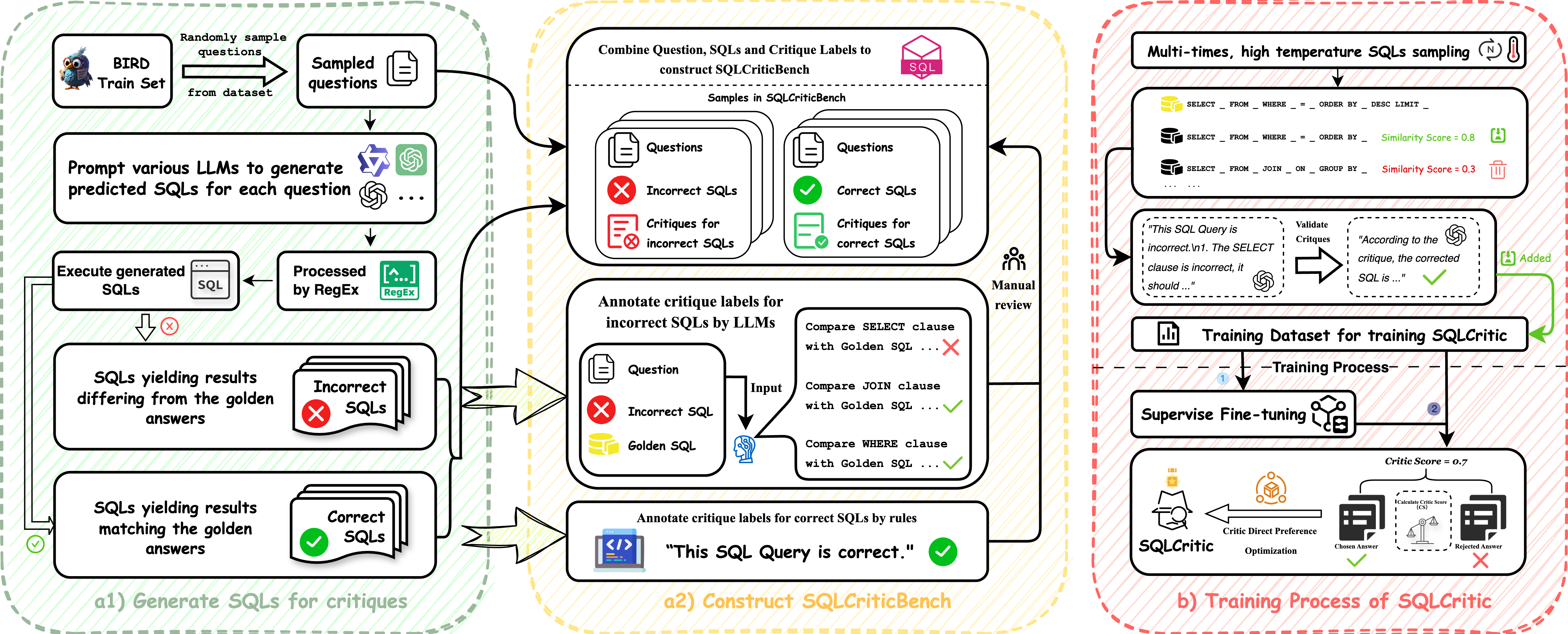}
    \caption{Overview of SQLCriticBench framework and SQLCritic training process (a1-a2: SQLCriticBench construction process, b: SQLCritic training process).}
    \label{fig:critic evaluation}
\end{figure*}

\section{Clause-wise Text-to-SQL Critique}
\label{Clause-wise-Text-to-SQL-Critique}
In this section, we first introduce the clause-wise text-to-sql critique generation task, which performs fine-grained error localization at the clause level and offers actionable feedback to facilitate SQL query corrections. 
We then present the training dataset curation process, followed by a detailed introduction of our SQLCritic model. 
The overall methodology is illustrated in Figure~\ref{fig:critic evaluation}.

\subsection{Task Definition}
\label{sec:task definition}
While correction-based evaluation offers a straightforward approach to validating the quality of generated critiques, it may lead to false positive errors—cases where an incorrect critique is mistakenly deemed correct simply because the resulting correction coincidentally produces the correct SQL query.
To address this, we propose the clause-wise text-to-sql critique evaluation task, which enables fine-grained evaluation of critique quality by assessing the accuracy of each critique component at clause level. 

Formally, given a natural language question $  Q  $ and a predicted SQL query $  \hat{S}  $ generated by a T2S model, the goal of the critique task is to produce a structured critique $C$, represented as a set of tuples:
$ C = \{(c_1, e_1), (c_2, e_2), \dots, (c_n, e_n)\} $
where each tuple $  (c_i, e_i)  $ is defined as follows:
\begin{itemize}
    \item $  c_i  $: the specific clause or component of $  \hat{S}  $ (e.g., SELECT, WHERE, GROUP BY) that diverges from the ground-truth SQL query $  S  $,
    \item $  e_i  $: a detailed explanation of the error in $  c_i  $, (e.g., The SELECT clause should select COUNT(*) instead of HEADS).
\end{itemize}
Compared with directly generating SQL critiques at the response level, the clause-wise critique output offers several advantages. 
First, it aligns more closely with the compositional structure of SQL, enabling precise error localization and clear interpretation. 
Second, it offers more actionable feedback, which can be more effectively used by large language models for error correction. 
Third, by decomposing critiques into clause-level components, this approach facilitates automatic evaluation by allowing more accurate comparison with human annotations.

\begin{figure}[t]
    \centering
    \includegraphics[width=1\linewidth]{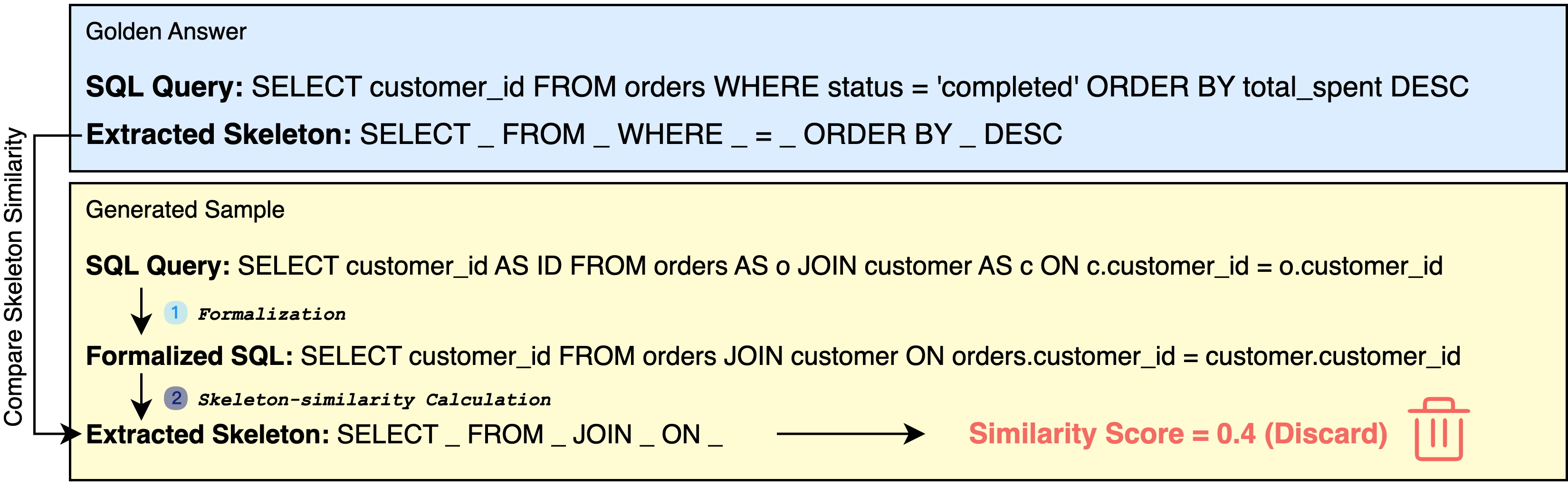}
    \caption{Illustration of filtering low-quality generated SQL query.}
    \label{fig:skeleton}
\end{figure}

\subsection{Training Dataset Curation} 
Before presenting our SQLCritic model for this clause-wise text-to-sql critique generation task, we first introduce the automatic training dataset curation process.

\noindent \textbf{Generating diverse SQL queries to be critiqued.} 
To comprehensively capture a broad spectrum of SQL query errors while maintaining a balanced level of correction difficulty, we employ two query generators: CodeS~\cite{li2024codes}, a state-of-the-art Text2SQL model, and Qwen2.5~\cite{qwen}, a general-purpose LLM with relatively lower performance on the Text2SQL task.
By using both models, CodeS introduces more high-quality and challenging-to-critique SQL queries, while Qwen2.5 includes more SQL queries that are easier to correct.
Specifically, we use BIRD~\cite{BIRD} as the data source and for each datapoint $(Q_i, S_i)$ in it, we apply \textit{high-temperature} and \textit{multi-sampling} strategies to generate the predicted SQL query $\hat{S}_i$. 
The use of \textit{high-temperature} and \textit{multi-sampling} further enriches the diversity of generated SQL queries, leading to a broader range of difficulty levels for critiquing and correcting.

\noindent \textbf{Filtering low-quality generated SQL queries.} 
However, upon generating the SQL queries, we observed that many of them contained trivial errors or were structurally misaligned with the golden labels, making them unsuitable for evaluation with current Text2SQL systems.
To address this issue, we adapt a two-stage process to filter low-quality SQL queries.
Specifically, we first normalize table and column names to lowercase and remove unnecessary elements such as aliases to avoid unwarranted superficial critique issues.
Next, we compute \textit{skeleton based similarity} to filter queries whose structural representations significantly deviate from the golden labels. 
To be more specific, we extract the skeleton representation of each query by replacing literals and standardizing table/column names with placeholder "\_". 
We then compute the Euclidean distance between the skeleton of each query and its corresponding gold-standard skeleton and discard queries with low similarity scores. 
The filtering process is illustrated in Figure {\ref{fig:skeleton}}.

By applying this filtering pipeline, we obtain a training dataset $ \mathcal{D} = \{(Q_i, S_i, \hat{S}_i)\}_{i=1}^N $ that is both diverse and focuses on more realistic Text2SQL errors.

\noindent \textbf{Annotating generated SQL queries with clause-wise critiques.} 
Given the collected dataset $ \mathcal{D} = \{(Q_i, S_i, \hat{S}_i)\}_{i=1}^N $, now we automatically annotate each SQL query $\hat{S}_i$ with clause-wise citique.
Specifically, we execute each generated SQL query and compare its answer with the gold-standard answer. 
Queries that produce correct results are annotated with the critique label: “This SQL query is correct.”
For incorrect SQL queries, we first use GPT-4 to compare each clause (e.g., SELECT, WHERE) of $\hat{S}_i$ with the corresponding clause of $S_i$ to generates clause-wise critiques $(c_i, e_i)$, where $c_i $ is the specific clause of $  \hat{S} $ and $  e_i  $ is the detailed explanation of the error in $  c_i  $.
Second, to validate the correctness of $(c_i, e_i)$, another LLM is used to correct the predicted SQL query $\hat{S}$ into a revised version $S'$ based solely on the clause-wise critiques $C=\{(c_i, e_i)\}|_{i=1}^{M}$ where $M$ is the total number of clauses. 
If the execution output of $S'$ eqauls to that of the ground-truth query, the corresponding critique is considered the ground-truth (preferred) critique label and is included in the training dataset.
An incorrect critique is also randomly sampled as the dispreferred counterpart, forming a critique pair for subsequent preference learning.
Finally, we automatically construct a clause-wise generation training dataset $ \mathcal{D}_\text{cirtic} = \{(Q_i, S_i, \hat{S}_i, C_i^w, C_i^l)\}_{i=1}^N $ where $C_i=\{(c^i_j, e^i_j)\}|_{j=1}^{M}$.

Notably, the clause-wise generation training dataset $\mathcal{D}$ includes both correct and erroneous samples, thereby challenging the critiquing model to effectively differentiate between accurate and flawed SQL queries.

\subsection{Clause-wise Critique Inconsistency Aware Direct Preference Optimization}
After constructing the high-quality training dataset $\mathcal{D}$, we propose a two-stage training process for training the SQLCritic model. 
First, we use Supervised Fine-Tuning (SFT)~\cite{sft} to equip the model with foundational critique capabilities. 
Subsequently, we propose a variant of DPO that is sensitive to clause-level inconsistencies between the preferred and dispreferred critiques, penalizing the critique model for generating substantial discrepancies in its clause-wise assessments.

\noindent \textbf{Warm-starting with Supervised Fine-tuning.} To warmup the learning process, the base model is first fine-tuned using supervised training to acquires an initial ciritique capability.

Specifically, given the training dataset $\mathcal{D}_\text{cirtic} = \{(Q_i, S_i, \hat{S}_i, C_i^w, C_i^l)\}_{i=1}^N $, the learning objective of SFT is as follows:
\begin{equation}
\label{eq:sft}
\resizebox{0.98\columnwidth}{!}{$
\mathcal{L}_{\text{SFT}}(\theta) = -\frac{1}{N} \sum_{i=1}^{N} \sum_{t=1}^{T_i} \log P_{\theta}(C^w_{i,t} \mid C^w_{i,<t}, Q^i, \hat{S}_i,)
$}
\end{equation}
where $T_i$ denotes the total number of tokens in $C_i^w$. 
$C^w_{i,t}$ refers to the $t$-th token of $C_i^w$, while $C^w_{i,<t}$ denotes the sequence of the first $t$ tokens. 

\paragraph{DPO tailored for clause-wise critique.}
In the second stage of training, we leverage preference learning\cite{rlhf} to further enhance the critique generation capabilities of the critique model.
In particular, we choose the offline preference optimization method Direct Preference Optimization (DPO)~\cite{dpo} as it is more stable and efficient compared to online RLHF methods\cite{rlhf}. 
The learning objective of DPO is directly formulated over the the policy model $\pi_\theta (y|x)$ and a reference model $\pi_\text{ref} (y|x)$:
\begin{equation}
    \begin{split}
   \mathcal{L}_\text{DPO}(\theta) = -\mathop{\mathbb{E}}_{(x, y_w, y_l) \in \mathcal{D}}\Bigl[\log & \sigma\Bigl(\beta\log \tfrac{\pi_\theta (y_w|x)}{\pi_{\text{ref}} (y_w|x)}\\ 
   &- \beta\log \tfrac{\pi_\theta (y_l|x)}{\pi_{\text{ref}} (y_l|x)}\Bigr)\Bigr]
    \end{split}
    \label{eq:dpo}
\end{equation}
where $y_l$ and $y_w$ are the rejected and chosen answer. $\sigma$ is the logistic function. $\beta$ is the hyper-parameter controlling the deviation from $\pi_\text{ref} (y|x)$. Action score $\log \pi(y|x)$ is the response generation likelihood.

The learning objective in Eq.~\ref{eq:dpo} uses a fixed $\beta$ value, however, \citet{wu2024beta} indicate that for preference pairs with smaller differences, the value of $\beta$ should be reduced to encourage larger updates. 
In contrast, for easily distinguishable pairs, the value of $\beta$ should be increased accordingly. 
Inspired by this insight, we propose a variant of DPO, CriticDPO, tailored for training the critique model, whose $\beta$ value is sensitive to clause-level inconsistencies between the preferred and dispreferred critiques.

Specifically, given the training dataset $ \mathcal{D}_\text{critic} = \{(Q_i, S_i, \hat{S}_i, C_i^w, C_i^l)\}_{i=1}^N $, the learning objective of the clause-wise critique inconsistency aware DPO is as follows:
\begin{equation}
\resizebox{0.95\columnwidth}{!}{$
    \begin{split}
   \mathcal{L}_\text{CriticDPO}(\theta) = -\mathop{\mathbb{E}}_{(Q_i, C_i^w, C_i^l) \in \mathcal{D}_\text{critic}}\Bigl[\log & \sigma \Bigl( S^i_{avg} \cdot \Bigl(\log \tfrac{\pi_\theta (C_i^w|Q_i)}{\pi_{\text{ref}} (C_i^w|Q_i)}\\ 
   &- \log \tfrac{\pi_\theta (C_i^l|Q_i)}{\pi_{\text{ref}} (C_i^l|Q_i)} \Bigr) \Bigr)\Bigr]
    \end{split}
$}
    \label{eq:critic_dpo}
\end{equation}
\begin{equation}
\resizebox{0.4\columnwidth}{!}{$
    S^i_{avg} = \beta + 0.1 * \text{CS}_i
$}
\end{equation}
where $\text{CS}_i$ is a dynamic coefficient designed to reflect the degree of clause-level judgment conflict between the preferred and dispreferred critiques within each training pair.

Specifically, $\text{CS}_i$ is calculated as follows.
We begin by evaluating whether the correctness judgments of the rejected and chosen critiques are consistent with respect to the criticized SQL query. 
If a conflict is detected—such as the rejected critique incorrectly labeling a correct SQL as erroneous, or vice versa—the critique is considered severely flawed, and $\text{CS}_i$ is set to -1 to reflect a large difference.
If the overall judgments align, we proceed to assess the fine-grained quality of the rejected critique. To be more specific, we calculate the averaged number of correctly identified clause-level critiques as $\text{CS}_i = \dfrac{1}{T}\sum_{j=1}^T c^i_j$ where $T$ is the total critique points in the chosen critique. 
$\text{CS}_i$ is subsequently scaled by a factor of 0.1 to normalize the score within the range [0, 0.1], which we found in our experiments is useful to mitigate large fluctuations during the training process. 
\section{Experiment}
\subsection{Experimental Setup}
\noindent \textbf{Datasets and Metrics.} 
We conduct experiments on critiquing and correcting SQL queries using the Spider \cite{spider} and BIRD \cite{BIRD} datasets, with Execution Accuracy (EX) and Valid Efficiency Score (VES) serving as our evaluation metrics. 
More details of these datasets can be seen in Appendix \ref{datasets}, and the definition of evaluation metrics is in Appendix \ref{metrics}.

\noindent \textbf{SQLCriticBench.}
To provide a more detailed assessment of the quality of critique texts generated by our SQLCritic model, we construct the SQLCriticBench benchmark, a comprehensive evaluation framework for Text-to-SQL critique tasks. SQLCriticBench consists of over 3000 samples, covering a diverse range of SQL queries along with their corresponding critique labels. The composition of the benchmark is summarized in Table~\ref{tab:SQLCriticBench}. In this benchmark, we evaluate the performance of our SQLCritic model against other powerful closed-source models such as GPT-3.5, GPT-4, and o1-mini~\cite{gpt4} in Text-to-SQL critique tasks. 
We use the Critique Performance Score (CPS) as the evaluation metric, which combines both classification accuracy and critique quality into a single interpretable metric.
The definition of CPS is detailed in Appendix \ref{app:evaluation of SQLCriticBench}.

\noindent \textbf{Baselines.} 
To study the impact of different refinement methods on Text-to-SQL performance, we compare our approach against several widely used correction strategies, including Self-Correction~\cite{dinsql}, Self-Consistency~\cite{dailsql}, Execution Feedback~\cite{macsql}, and MAGIC~\cite{magic}. Detailed descriptions of these baselines are provided in Appendix~\ref{baselines}. Additionally, for the evaluation of critique ability, we use strong closed-source models such as GPT-4 as baselines.
\begin{table}[t]
\resizebox{\columnwidth}{!}{%
\begin{tabular}{@{}cccc@{}}
\toprule
Model &
  \begin{tabular}[c]{@{}c@{}}Negative / Positive \\ Samples\end{tabular} &
  \begin{tabular}[c]{@{}c@{}}Avg. Incorrect \\ Clauses\end{tabular} &
  \begin{tabular}[c]{@{}c@{}}Avg. Length \\ of Critique\end{tabular} \\ \midrule
CodeS   & 300 / 685 & 3.05 & 94.89 \\ \midrule
GPT-3.5 & 100 / 572 & 3.97 & 138.62 \\ \midrule
GPT-4o  & 100 / 598 & 4.09 & 208.20 \\ \midrule
o1-mini & 100 / 564 & 3.57 & 138.30 \\ \midrule
Total / Avg.   & 3019 & 3.67 & 145.00 \\ \bottomrule
\end{tabular}%
}
\caption{The statistical distribution of SQLCriticBench. Avg. represents the average value. Total / Avg. represents that the value in this column is either the total (for the first column) or the average (for the other columns).}
\label{tab:SQLCriticBench}
\end{table}

\begin{table*}[h]
\centering
\resizebox{\textwidth}{!}{%
\begin{tabular}{c|cccccccccccccc}
\toprule
\textbf{Model} & & & \multicolumn{4}{c}{\textbf{Correcting Incorrect SQLs}} &  \multicolumn{4}{c}{\textbf{SQLs with Exec Error}} & \multicolumn{4}{c}{\textbf{All SQLs}}\\
\cmidrule(r){4-7} \cmidrule(r){8-11} \cmidrule(r){12-15} 
 & &  & \small \textbf{EX (\%)} &  &  & \small \textbf{VES (\%)}  & \small \textbf{EX (\%)} &  &  & \small \textbf{VES (\%)} & \small \textbf{EX (\%)} &  &  & \small \textbf{VES (\%)} \\
\midrule
\multicolumn{1}{l}{{\textbf {GPT-3.5-turbo (baseline)}}} & & & \small 41.98 &  &  & \small 47.78 & \small 41.98 &  &  & \small 47.78 & \small 41.98 &  &  & \small 47.78\\
\midrule

w/ Self-Correction (DIN-SQL)     & & & \small 45.70 $\uparrow$ &  &  & \small 47.51 & \small 43.55 $\uparrow$ &  &  & \small 45.58 & \small 41.07 &  &  & \small 41.79\\
w/ Self-Consistency(DAIL-SQL)     & & & \small 50.98 $\uparrow$ &  &  & \small 52.90 $\uparrow$ & \small 46.61 $\uparrow$ &  &  & \small 48.60 & \small 46.35 $\uparrow$ &  &  & \small 47.28 $\uparrow$\\
w/ Execution Feedback (MAC-SQL)     & & & \small 49.44 $\uparrow$ &  &  & \small 49.44 $\uparrow$ & \small 46.54 $\uparrow$ &  &  & \small 49.44 $\uparrow$ & \small 46.54 $\uparrow$ &  &  & \small 49.44 $\uparrow$\\
w/ Auto-Generated Guideline (MAGIC)     & & & \small 45.18 $\uparrow$ &  &  & \small 47.13 & \small 44.07 $\uparrow$ &  &  & \small 45.95 & \small 43.61 $\uparrow$ &  &  & \small 44.49\\
w/ \textbf{SQLCritic}     & & & \small \textbf{51.39} $\uparrow$ &  &  & \small \textbf{53.22} $\uparrow$ & \small \textbf{46.54} $\uparrow$ &  &  & \small \textbf{49.44} $\uparrow$ & \small \textbf{48.94} $\uparrow$ &  &  & \small \textbf{51.32} $\uparrow$\\

\midrule
\multicolumn{1}{l}{{\textbf {GPT-4o-mini (baseline)}}} & & & \small 39.77 &  &  & \small 40.22 & \small 39.77 &  &  & \small 40.22 & \small 39.77 &  &  & \small 40.22 \\
\midrule

w/ Self-Correction (DIN-SQL)     & & & \small 43.48 &  &  & \small 44.71 & \small 40.94 &  &  & \small 41.65 & \small 39.18 &  &  & \small 40.23$\uparrow$\\
w/ Self-Consistency(DAIL-SQL)    & & & \small  39.77 &  &  & \small 40.32 & \small  39.77 &  &  & \small 40.33 & \small 39.77 &  &  & \small 40.24$\uparrow$\\
w/ Execution Feedback (MAC-SQL)     & & & \small 43.29$\uparrow$ &  &  & \small 43.87$\uparrow$ & \small 43.29$\uparrow$ &  &  & \small 43.87$\uparrow$ & \small 43.29$\uparrow$ &  &  & \small 43.87$\uparrow$\\
w/ \textbf{SQLCritic}     & & & \small \textbf{48.32}$\uparrow$ &  &  & \small \textbf{49.90}$\uparrow$ & \small \textbf{44.29}$\uparrow$ &  &  & \small \textbf{44.87}$\uparrow$ & \small \textbf{47.79}$\uparrow$ &  &  & \small \textbf{49.21}$\uparrow$\\

\midrule
\multicolumn{1}{l}{{\textbf {o1-mini (baseline)}}} & & & \small 47.85 &  &  & \small 49.66 & \small 47.85 &  &  & \small 49.66 & \small 47.85 &  &  & \small 49.66 \\
\midrule

w/ Self-Correction (DIN-SQL)     & & & \small 48.17$\uparrow$ &  &  & \small 50.01$\uparrow$ & \small 48.17$\uparrow$ &  &  & \small 50.18$\uparrow$ & \small 45.76 &  &  & \small 47.55\\
w/ Self-Consistency(DAIL-SQL)     & & & \small 50.13$\uparrow$ &  &  & \small 51.94$\uparrow$ & \small  48.11$\uparrow$ &  &  & \small 49.97$\uparrow$ & \small 48.31$\uparrow$ &  &  & \small 50.13$\uparrow$\\
w/ Execution Feedback (MAC-SQL)     & & & \small 50.78$\uparrow$ &  &  & \small 52.93$\uparrow$ & \small 50.78$\uparrow$ &  &  & \small 52.93$\uparrow$ & \small 50.78$\uparrow$ &  &  & \small 52.93$\uparrow$\\
w/ \textbf{SQLCritic}     & & & \small \textbf{54.78}$\uparrow$ &  &  & \small \textbf{57.12}$\uparrow$ & \small \textbf{50.78}$\uparrow$ &  &  & \small \textbf{52.93}$\uparrow$ & \small \textbf{52.14}$\uparrow$ &  &  & \small \textbf{53.32}$\uparrow$\\

\midrule
\multicolumn{1}{l}{{\textbf {GPT-4o (baseline)}}} & & & \small 49.87 &  &  & \small 52.43 & \small 49.87 &  &  & \small 52.43 & \small 49.87 &  &  & \small 52.43 \\
\midrule

w/ Self-Correction (DIN-SQL)    & & & \small 49.67 &  &  & \small 52.22 & \small 49.87 &  &  & \small 52.30 $\uparrow$ & \small 47.72 &  &  & \small 50.26\\
w/ Self-Consistency(DAIL-SQL)    & & & \small 54.17$\uparrow$ &  &  & \small 57.40$\uparrow$ & \small 51.50$\uparrow$ &  &  & \small 53.93$\uparrow$ & \small 46.81 &  &  & \small 48.76\\
w/ Execution Feedback (MAC-SQL)    & & & \small 52.80 $\uparrow$ &  &  & \small 55.32 $\uparrow$ & \small 52.80 $\uparrow$ &  &  & \small 55.32 $\uparrow$ & \small 52.80 $\uparrow$ &  &  & \small 55.32 $\uparrow$\\
w/ \textbf{SQLCritic}     & & & \small \textbf{55.82}$\uparrow$ &  &  & \small \textbf{57.64}$\uparrow$ & \small \textbf{52.80}$\uparrow$ &  &  & \small \textbf{55.32}$\uparrow$ & \small \textbf{54.58} $\uparrow$ &  &  & \small \textbf{55.98} $\uparrow$\\

\midrule
\multicolumn{1}{l}{{\textbf {CodeS-7B (baseline)}}} & & & \small 57.17 &  &  & \small 58.80  & \small 57.17  &  &  & \small 58.80  & \small 57.17 &  &  & \small 58.80\\
\midrule

w/ Self-Correction (DIN-SQL)    & & & \small 56.84 &  &  & \small 60.37$\uparrow$ & \small 57.11 &  &  & \small 60.64$\uparrow$ & \small 53.98 &  &  & \small 57.26\\
w/ Self-Consistency(DAIL-SQL)    & & & \small 59.00$\uparrow$ &  &  & \small 62.34$\uparrow$ & \small 57.63$\uparrow$ &  &  & \small 61.09$\uparrow$ & \small 57.69$\uparrow$ &  &  & \small 60.78$\uparrow$\\
w/ Execution Feedback (MAC-SQL)    & & & \small {59.00} $\uparrow$ &  &  & \small 63.30 $\uparrow$ & \small 59.00 $\uparrow$ &  &  & \small 63.30 $\uparrow$ & \small 59.00 $\uparrow$ &  &  & \small 63.30 $\uparrow$\\
w/ \textbf{SQLCritic}     & & & \small \textbf{59.45}$\uparrow$ &  &  & \small \textbf{63.47}$\uparrow$ & \small \textbf{59.00}$\uparrow$ &  &  & \small \textbf{63.30}$\uparrow$ & \small \textbf{59.34} $\uparrow$ &  &  & \small \textbf{64.28} $\uparrow$\\

\bottomrule
\end{tabular}%
}
\caption{Main results on the BIRD benchmark. An upward arrow ($ \uparrow $) indicates an improvement over the baseline, and bold values represent the best results.}
\label{tab:main result}
\end{table*}

\begin{table}[t]
\centering
\setlength{\abovecaptionskip}{0.2cm}
\setlength{\belowcaptionskip}{-0.4cm}
\setlength{\tabcolsep}{3pt}
\resizebox{\columnwidth}{!}{
\begin{tabular}{@{}ccccc@{}}
\toprule
Model &
  \begin{tabular}[c]{@{}c@{}}Fail in \\ Error Detection\end{tabular} &
  \begin{tabular}[c]{@{}c@{}}Flaw in \\ textual critique\end{tabular} &
  \begin{tabular}[c]{@{}c@{}}Correct in \\ both stages\end{tabular} &
  \begin{tabular}[c]{@{}c@{}}Final Score\end{tabular} \\ \midrule
GPT-3.5   & 1257 & 278 & 1484 & 5.4 / 100  \\ \midrule
GPT-4o    & 1195 & 273 & 1551 & 12.1 / 100 \\ \midrule
o1-mini   & 1186 & 275 & 1558 & 12.3 / 100 \\ \midrule
GPT-4     & 955  & 259 & 1805 & 16.8 / 100 \\ \midrule
SQLCritic & \textbf{684}  & \textbf{256} & \textbf{2079} & \textbf{48.4} / 100 \\ \bottomrule
\end{tabular}%
}
\caption{The critique ability evaluation on SQLCriticBench. Bold values represent the best results.}
\label{tab:critique_eval}
\end{table}

\begin{table}[t]
\resizebox{\columnwidth}{!}{%
\begin{tabular}{lcccccc}
\toprule
\textbf{Model} & & & \multicolumn{4}{c}{\textbf{All SQLs}}\\
\cmidrule(r){4-7} 
 & &  & \small \textbf{EX (\%)} &  &  & \small \textbf{VES (\%)} \\
\midrule

\multicolumn{1}{l}{\small {\textbf {GPT-3.5-turbo}}} & & & \small 41.98 &  &  & \small 47.78\\
\midrule

w/ Holistic Critique    & & & \small 37.72 & & & \small 43.38 \\
w/ Clause-wise Critique     & & & \small 40.72 &  &  & \small 45.83 \\
w/ SQLCritic     & & &  \small 48.94  &  &  & \small 51.32  \\

\midrule
\multicolumn{1}{l}{\small {\textbf {GPT-4o}}} & & & \small 49.87 &  &  & \small 52.43\\
\midrule

w/ Holistic Critique    & & & \small 45.12 & & & \small 48.36 \\
w/ Clause-wise Critique     & & & \small 47.97 &  &  & \small 50.03 \\
w/ SQLCritic     & & &  \small 54.58  &  &  & \small 55.98  \\

\bottomrule
\end{tabular}%
}
\caption{Correction performance under different sets of critique methods. w/ Holistic Critique and w/ Clause-wise Critique refer to using critiques in their respective formats, generated by LLMs using prompts. w/ SQLCritic refers to using critiques provided by our trained SQLCritic model.}
\label{tab:clause-wise critique}
\end{table}
\noindent\textbf{Implementation Details.} We train Qwen-2.5-14B using 4 A100 (80G) GPUs. 
The SFT stage is conducted for 3 epochs with a learning rate of 2e-5. 
This is followed by Critic DPO training for 1 epoch with a learning rate of 2e-6.

\subsection{Critiquing and Correcting Results on BIRD and Spider}
\label{sec:main result}

\begin{table}[t]
\centering
\setlength{\abovecaptionskip}{0.2cm}
\setlength{\belowcaptionskip}{-0.4cm}
\footnotesize
\resizebox{\columnwidth}{!}{%
\begin{tabular}{lcccccc}
\toprule
\textbf{Model} & & & \multicolumn{4}{c}{\textbf{All SQLs}}\\
\cmidrule(r){4-7} 
 & &  & \small \textbf{EX (\%)} &  &  & \small \textbf{VES (\%)} \\






\midrule
\multicolumn{1}{l}{\small {\textbf {GPT-4o}}} & & & \small 49.87 &  &  & \small 52.43\\
\midrule

w/ SFT + Critic DPO    & & & \small \textbf{54.58} & & & \small \textbf{55.98} \\
w/ SFT + DPO     & & & \small 52.80 &  &  & \small 54.54 \\
w/ SFT     & & &  \small 51.92  &  &  & \small 53.34  \\

\midrule
\multicolumn{1}{l}{\small {\textbf {CodeS-7B}}} & & & \small 57.17 &  &  & \small 58.80\\
\midrule

w/ SFT + Critic DPO    & & & \small \textbf{59.34} & & & \small \textbf{64.28} \\
w/ SFT + DPO     & & & \small 59.19 &  &  & \small 63.26 \\
w/ SFT     & & &  \small 57.28  &  &  & \small 60.97  \\
\bottomrule
\end{tabular}%
}
\caption{Ablation study on correction performance using critiques from SQLCritic trained with different methods. Bold represents the best result.}
\label{tab:ablation study}
\end{table}

To evaluate the effectiveness of SQLCritic, we design three experimental settings to explore its performance across different scenarios: (1) correcting manually identified incorrect SQL statements, (2) addressing SQL queries that produce explicit execution errors, and (3) handling all SQL outputs to simulate real-world conditions. The evaluation is conducted on model-generated SQLs from the BIRD and Spider datasets \cite{spider}, where SQLCritic iteratively guides GPT-3.5-turbo to refine flawed queries and align them with user intent.

Experimental results in Table~\ref{tab:main result} show that previous methods, such as self-correction and self-consistency, excel in controlled settings with manually identified incorrect SQLs. However, their performance declines in real-world scenarios, where models must independently identify errors; self-correction often modifies correct queries, while self-consistency shows only marginal gains.
In contrast, SQLCritic performs robustly across all settings and consistently outperforms other baseline methods. It generates precise critiques, enabling effective SQL revision in controlled experiments, and excels in real-world conditions by accurately identifying errors and guiding corrections, significantly improving SQL generation and correction quality. We also conduct experiment on the Spider dataset \cite{spider} in Appendix \ref{app:spider}, and SQLCritic still achieves substantial improvements across all correction methods.

\subsection{Clause-wise Critique Evaluation on SQLCritcBench}
\label{sec:critique eval}
As shown in Table~\ref{tab:critique_eval}, the evaluation framework includes four key metrics to comprehensively assess critique performance: \textbf{Fail in Error Detection} measures the number of instances where the model fails to correctly classify whether an SQL prediction is accurate or not; \textbf{Flaw in Textual Critique} evaluates the number of inaccurate critique content, reflecting the model's ability to provide accurate and useful explanations to guide the refinement; \textbf{Correct in Both Stages} represents the number of instances where the model succeed in both error detection and textual critique tasks; and \textbf{Final Score (CPS)} aggregates the model's overall performance into a normalized score ranging from 0 to 100.

The results demonstrate SQLCritic’s significant superiority over other baseline models. It achieves the lowest classification errors (684), the fewest critique flaws (256), and the highest number of correct evaluations across both stages (2079), yielding the best CPS score of 48.4. This highlights SQLCritic’s robust capabilities in both \textbf{error identification} and \textbf{textual critique generation}. In contrast, other models exhibit notable limitations: GPT-3.5 and GPT-4o struggle with substantially higher classification errors (1257 and 1195, respectively), while GPT-4, though showing marginal improvement, attains only a CPS score of 16.8, indicating persistent weaknesses in generating high-quality textual critiques. These findings collectively underscore SQLCritic’s clear performance advantage in the Text-to-SQL critique task.

\subsection{Ablation Studies}
\noindent\textbf{Comparing Clause-wise Critique with Holistic Critique.}
This ablation study compares \textbf{clause-wise critique} and \textbf{holistic critique} in SQL correction tasks, focusing on performance and stability. Holistic critique corrects the entire SQL query in a single step, while clause-wise critique targets specific clauses for granular adjustments. We apply both critique strategies to refine GPT-3.5-turbo and GPT-4o predictions \cite{gpt4o} on the BIRD development set. As shown in Table~\ref{tab:clause-wise critique}, holistic critique lead to a noticeable performance decline, whereas clause-wise critique result in only minor degradation. The decline in both cases may stem from the inherent limitations of self-correction methods, which can struggle with compounding errors. However, clause-wise critique still demonstrated greater stability and effectiveness. Furthermore, employing the SQLCritic model significantly improves performance, further validating the efficacy and superiority of our method.  


\paragraph{Comparing SFT, DPO and Critic DPO.}  
We conduct an ablation study to evaluate the effectiveness of the proposed Critic DPO method in optimizing the SQLCritic model’s clause-wise critique performance. Specifically, we compare three training variations: SFT, SFT with DPO, and SFT with Critic DPO. Results in Table \ref{tab:ablation study} demonstrate that SFT + Critic DPO outperforms the other approaches on both EX and VES evaluation metrics. This improvement arises from the adaptive, clause-level optimization process in Critic DPO, which utilizes clause-wise critiques to discern subtle preference differences, thereby achieving more precise and effective model alignment.

\subsection{Error Analysis}
\label{sec:error analysis}

\begin{table}[t]
\resizebox{\columnwidth}{!}{%
\begin{tabular}{@{}cccc@{}}
\toprule
Alignment Type & \begin{tabular}[c]{@{}c@{}}Align with\\ Huamn\end{tabular} & \begin{tabular}[c]{@{}c@{}}Doesn't Align\\ with Huamn\end{tabular} & Acc  \\ \midrule
Number         & 72 & 28 & 72\% \\ \bottomrule
\end{tabular}%
}
\caption{The Human Consistency Experiment}
\label{tab:human alignment}
\end{table}

\begin{figure}[t]
    \centering
    \includegraphics[width=0.9\linewidth]{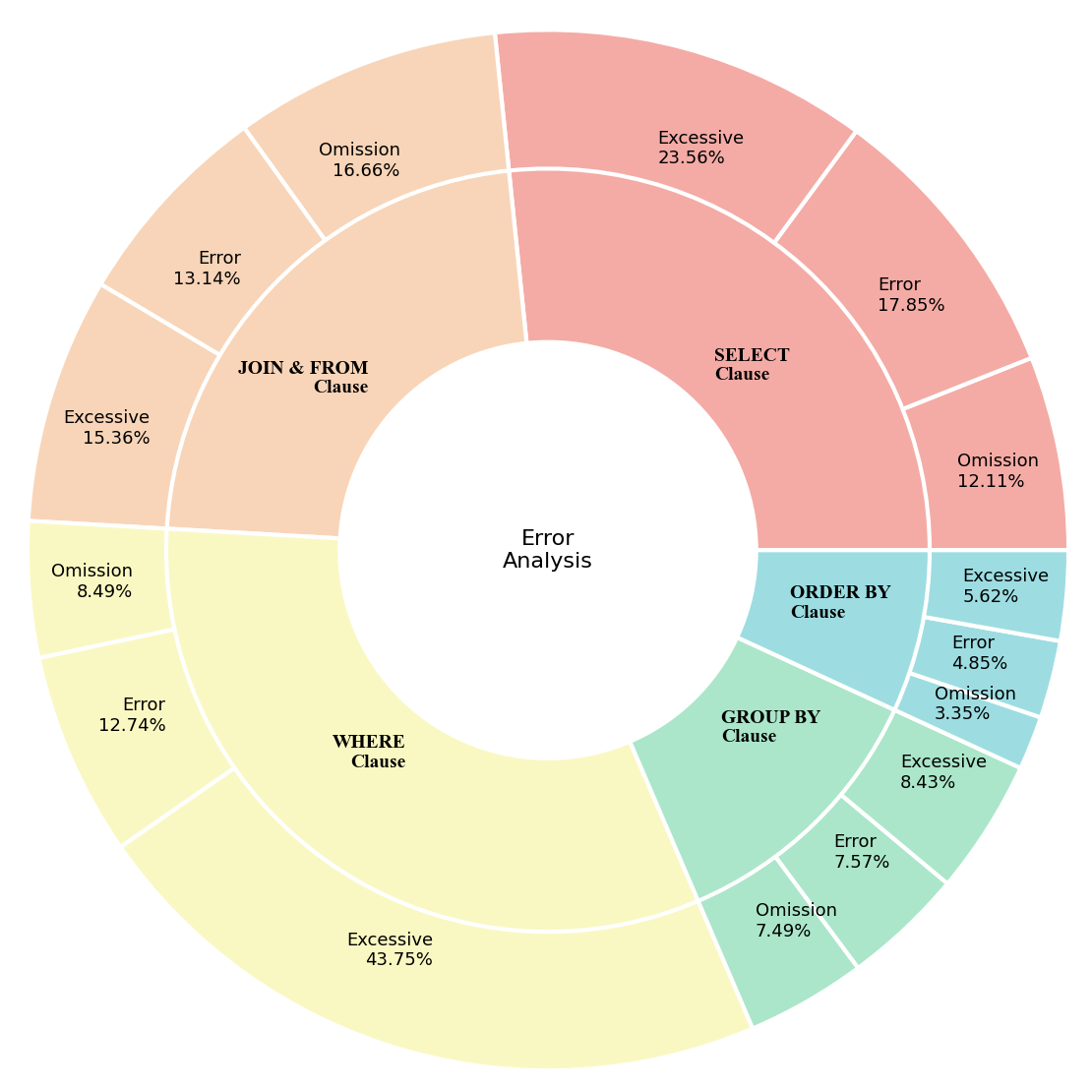}
    \caption{Error analysis distribution. Different colors represent the distribution of different clauses.}
    \label{fig:error_analysis}
\end{figure}

We conduct a detailed error analysis of SQLCritic's critique performance on our SQLCriticBench. Specifically, we randomly select 200 samples from SQLCriticBench and analyze SQLCritic's performance across individual SQL clauses. For each clause, we categorize errors into three types: \textbf{Critique Omission}, where the model fails to identify and critique erroneous SQL clauses; \textbf{Critique Error}, involving incorrect or logically invalid feedback for flawed clauses; and \textbf{Excessive Critique}, characterized by unwarranted flagging of valid clauses as erroneous.

As shown in Figure~\ref{fig:error_analysis}, SQLCritic’s performance varies significantly across SQL clauses. While the model demonstrates strong general proficiency in identifying errors, persistent challenges arise in nuanced scenarios involving complex logic or semantic ambiguity. The WHERE clause exhibits the highest critique error rate (43.75\%) primarily due to difficulties in parsing nested logical conditions, implicit semantic errors, and context-dependent constraints. Similarly, the critique error rate for SELECT clauses has reached 23.56\%, which may be due to the diversity of selection criteria, ambiguous column references, and edge cases involving computations or aggregations. Critique omissions and excessive critiques in the JOIN and FROM clauses further highlight limitations in modeling table relationships and schema dependencies. Therefore, future optimizations should focus on improving table alignment understanding and reducing unnecessary critiques. Despite SQLCritic’s strengths, significant room remains for enhancing accuracy and precision in feedback generation, highlighting the inherent complexity of Text-to-SQL critique tasks.

\section{Human Consistency Experiments on CPS Evaluation Metrics}
While certain experimental findings \cite{rrhf} suggest that advanced LLMs, such as GPT-4, can attain high consistency with human evaluations, accurately measuring the performance of generative models nonetheless depends on analyzing their alignment with human-labeled data.

To evaluate the CPS metric, we randomly selected 100 samples from the SQLCriticBench. Each sample was independently reviewed by an expert, who determined whether the judgment given by GPT is consistent with human expert. Subsequently, the accuracy of CPS scored by model was calculated to measure its consistency with the expert's assessments. As shown in table \ref{tab:human alignment}, the CPS metric exhibits high agreement with human evaluations, demonstrating its effectiveness in assessing the critique quality.

\section{Conclusion}
In this work, we introduced SQLCritic and SQLCriticBench for Text-to-SQL error correction. 
By employing clause-wise critique, SQLCritic effectively localizes both syntactic and semantic errors in SQLs, and provide actionable and helpful critique to guide correction. 
Complementing this, SQLCriticBench provides a fine-grained evaluation benchmark with over 3,000 samples, addressing critical gaps in existing datasets of Text-to-SQL critique domain. 
Experiments demonstrated that SQLCritic significantly improves SQL accuracy on the BIRD and Spider datasets, and the results on SQLCriticBench further reveals its superior critique capabilities compared to existing models. 

\section*{Limitations}





While SQLCritic shows strength in clause-wise critiques, some areas for improvement exist. The model may struggle with deeply nested subqueries or complex conditionals (e.g., \textit{WHERE id IN (SELECT ...) AND (CASE...END)}), since breaking them down may overlook subtle semantic nuances. Its focus on individual clauses can also miss cross-clause dependencies, such as mismatches between \textit{GROUP BY} and \textit{HAVING} clauses or inconsistencies between \textit{SELECT} columns and \textit{JOIN} conditions, which a broader analysis could address. Performance also depends on training data diversity, with rare or unconventional SQL patterns potentially affecting real-world reliability. Addressing these areas could involve refining decomposition techniques, integrating cross-clause analysis, and diversifying training data to enhance robustness without compromising existing strengths.

\bibliography{acl_latex}

\clearpage

\begin{appendices}  

\section{Evaluation Metrics}
\label{metrics}

\subsection{Metrics for Critiquing and Correcting Evaluation}
We employ two key metrics to evaluate the performance of SQL correction result:

\begin{itemize}
[leftmargin=*,itemsep=1pt]
    \item \textbf{Execution Accuracy (EX):} Execution Accuracy measures the correctness of generated SQL queries by comparing their execution results with those of the ground truth. A generated query is considered correct if its execution output matches the reference query. \cite{qin2022surveytexttosqlparsingconcepts}

    \item \textbf{Valid Efficiency Score (VES):} The Valid Efficiency Score evaluates the execution efficiency of generated SQL queries by focusing on their accuracy and execution time. This metric considers only queries whose execution results align with the ground truth, ensuring a fair assessment of both accuracy and efficiency.
\end{itemize}

\subsection{Metric for Critique Ability Evaluation}
\label{app:evaluation of SQLCriticBench}
We believe that a high-quality critique should encompass the following two key capabilities: (1) \textbf{Accuracy in Error Detection}: The model's ability to effectively distinguish whether a given sample requires criticism; (2) \textbf{Quality of the Generated Critique}: For samples that do require criticism, the model's ability to precisely and comprehensively identify the errors, ensuring that the critique is both complete (without omissions) and concise (without redundancy). 

Thus, in order to accurately evaluate a model's critique ability. We propose the \textbf{Critique Performance Score (CPS)}, which combines both classification accuracy and critique quality into a single interpretable metric. CPS is defined as:

\begin{equation}
\text{CPS} = \frac{1}{N} \sum_{i=1}^N 
\mathbb{I}(\hat{y}_{bc}^i = y_{bc}^i) 
\cdot 
CQ
\end{equation}

To evaluate the quality of generated critiques, We first calculate $\mathbb{I}(\hat{y}_{bc}^i = y_{bc}^i)$ to reflect the accuracy of error detection, $\hat{y}_{bc}^i$ means the ground truth of error detection result while $y_{bc}^i$ indicates the predicted error detection result. Then, for the \textbf{Critique Quality Score(CQ)}, we use GPT-based scoring to compare the model-generated critique \(c\) with the human-labeled ground truth \(\hat{c}\). It's shown by existing work, LLMs can reliably evaluate free-form text given well-defined evaluation criteria \cite{criticbench}. We use GPT \cite{gpt4} to evaluate the critiques generated by the model against the annotated label critique on a clause-by-clause basis. For each clause, we classify it into one of four categories:

\begin{itemize}
    \item \textbf{Exact Match}: The clause predicted by the model appears in the label critique, and the critique content is completely consistent.
    \item \textbf{Partial Match}: The clause predicted by the model appears in the label critique, but the critique content is only partially consistent.
    \item \textbf{Error Match}: The clause predicted by the model appears in the label critique, but the critique content is inconsistent with the label critique.
    \item \textbf{Redundant}: A clause predicted by the model is unnecessary to critique, yet it appears in the predicted critique.
\end{itemize}

For different evaluation outcomes, we assign different scores: an Exact Match is scored as 1, a Partial Match is scored as 0.5, an Error Match is scored as 0, and a Redundant is scored as -0.3. After aggregating the scores for all clauses in a critique, we normalize the total score by dividing it by the number of critique points in the label critique. This yields a score between 0 and 1, which we refer to as the CQ value of one critique text.

The updated CPS metric ensures that critique models are evaluated holistically, reflecting their ability to both identify critique-worthy cases and generate high-quality feedback that aligns closely with human ground truth.

\section{SQLCritic on Spider}
\label{app:spider}

\begin{table}[t] 
\centering
\resizebox{\linewidth}{!}{ 
\begin{tabular}{@{}ccccccccc@{}}
\toprule
& \multicolumn{4}{c}{Before Correction}     & \multicolumn{4}{c}{After Correction}\\ 
\midrule
& GPT-3.5-turbo & GPT-4 & GPT-4o & CodeS-7B & GPT-3.5-turbo & GPT-4 & GPT-4o & CodeS-7B\\ 
\midrule
Easy   & 87.9\%         & 87.1\% & 89.1\%  & 91.9\%    & 89.1\%         & 89.9\% & 89.9\%  & 94.8\%    \\ \midrule
Medium & 78.5\%         & 82.7\% & 80.0\%  & 89.2\%    & 78.7\%         & 84.3\% & 80.9\%  & 90.6\%    \\ \midrule
Hard   & 60.9\%         & 72.4\% & 65.5\%  & 73.0\%    & 60.9\%         & 75.3\% & 65.5\%  & 73.6\%    \\ \midrule
Extra  & 47.6\%         & 50.6\% & 52.4\%  & 65.7\%    & 49.4\%         & 53.0\% & 52.4\%  & 65.1\%    \\ \midrule
All    & 72.8\%         & 76.9\% & 75.3\%  & 83.4\%    & 73.5\%         & 79.1\% & 75.9\%  & 84.6\%    \\ \bottomrule
\end{tabular}}
\caption{The Main Result on Spider.}
\label{tab:spider}
\end{table}

We applied our trained SQLCritic to critique the predictions of various open-source and proprietary models, including GPT-3.5-turbo, GPT-4, GPT-4o, and CodeS-7B, on the Spider dataset. Subsequently, we utilized GPT-3.5 to refine the predictions based on the provided critiques generated by our SQLCritic model. Experimental results demonstrate that the process significantly improves the prediction performance across all models. The  result is reported in Table \ref{tab:spider}

\section{Construction of SQLCriticBench}
\label{app:construction of SQLCriticBench}
To conduct a more comprehensive performance analysis of our SQLCritic model, there is an urgent need for a dedicated dataset specifically designed for the Text-to-SQL critique domain. Currently, there is no benchmark specifically designed to assess model capabilities in performing the Text-to-SQL critique generation task as defined in Section \ref{sec:task definition}. This gap leaves current evaluation methods unable to measure a model’s ability to identify fine-grained issues (e.g., syntax errors and semantic errors) in SQL. To address this limitation, we introduce \textbf{SQLCriticBench}, a benchmark dataset comprising over 3000 samples, each sample is a set of SQL outputs generated by diverse models alongside human-verified clause-wise critiques, offering a comprehensive and challenging test suite to evaluate the performance of critique generation models. 

As shown in Figure \ref{fig:critic evaluation}, the construction of SQLCriticBench includes two main parts: (1) Generate SQLs for critiques (Figure \ref{fig:critic evaluation} a1) and (2) Label critiques for SQLs (Figure \ref{fig:critic evaluation} a2).

\paragraph{Generate SQLs for Critiques:}

In the first step, we prompt several state-of-the-art SQL generation models (e.g., GPT-4 \cite{gpt4}, Qwen2.5 \cite{qwen}) to produce predicted SQL queries based on the BIRD \cite{BIRD} dataset. These predicted SQL queries often exhibit a range of errors, such as missing clauses, mismatched logical conditions, or incorrect aggregation operations. For each predicted SQL, we compare it with the corresponding gold SQL to classify the predictions into two categories: Correct SQLs and Incorrect SQLs.

\paragraph{Label Critiques for SQLs:}

Correct SQLs are directly labeled with the statement: "This SQL query is correct." For incorrect SQLs, clause-wise critiques are generated as labels to identify and explain the discrepancies. Specifically, we use GPT-4 \cite{gpt4} with a carefully designed prompt, where each incorrect SQL and its corresponding Gold SQL pair is provided as input. The prompt explicitly instructs GPT-4 to:

\begin{enumerate}
\item Compare each clause in the incorrect SQL with those in the Gold SQL (e.g., SELECT, WHERE, GROUP BY, etc.);
\item Identify clauses that contain errors or inconsistencies;
\item Generate detailed critiques describing the semantic errors, explaining the causes of the mistakes, and suggesting potential corrections.
\end{enumerate}

To ensure the quality and reliability of the GPT-4-generated critiques, all critique labels undergo a thorough manual review process. Two independent reviewers assess each critique to verify that all syntax and semantic issues have been correctly identified and sufficiently addressed. Common manual adjustments include refining error descriptions, supplementing missing critique details, and enhancing the clarity and precision of the feedback.
 
Ultimately, the SQLCriticBench aggregates over 3000 collection of SQL predictions and critique labels, covering various error types from multiple models.

\section{Datasets}
\label{datasets}
\begin{itemize}
    \item \textbf{Spider} \cite{spider}: Spider is a benchmark dataset for complex SQL generation, containing 10,181 questions and 5,693 unique SQL queries across 200 databases from 138 domains. The datasets are split into non-overlapping subsets for training, development, and testing. SQL queries in Spider vary across four difficulty levels, making it ideal for evaluating models on multi-table reasoning and diverse query complexities.
    \item \textbf{BIRD} \cite{BIRD}: BIRD is a large-scale dataset comprising 12,751 question-SQL pairs across 95 databases from 37 professional domains like healthcare and blockchain. It includes external knowledge (e.g., numeric reasoning, domain knowledge) and sample table rows to aid schema linking. BIRD’s SQL queries are more complex than Spider’s, offering unique challenges for models handling intricate queries and external knowledge integration.
    \item \textbf{SQLCriticBench}: SQLCriticBench is a specialized benchmark for evaluating models on SQL critique tasks. It contains over 3,000 samples, each consisting of a question $Q$, a predicted SQL query $\hat{S}$, and a critique label $C$. The dataset is designed to assess a model’s ability to detect and provide feedback on errors in SQL queries, making it a valuable resource for improving SQL generation and critique performance.
\end{itemize}

\section{Baselines}
\label{baselines}
We compared our proposed method against the following widely-used SQL correction strategies:

\begin{itemize}
    \item \textbf{Self-Correction (DIN-SQL)} \cite{dinsql}: This method uses prompts to guide the model in self-correction. The model identifies errors in its generated SQL statements and iteratively corrects them.
    \item \textbf{Self-Consistency (DAIL-SQL)} \cite{dailsql}: Multiple candidate SQL statements are generated during inference. A voting mechanism selects the most frequent candidate as the final output. If no executable candidate is found, the method falls back to the first executable SQL or the first generated SQL.
    \item \textbf{Execution Feedback (MAC-SQL)} \cite{macsql}: By leveraging explicit feedback from the SQL executor, such as syntax error messages, this method incorporates error information into correction prompts to guide further improvements.
    \item \textbf{MAGIC} \cite{magic}: A specialized SQL correction framework that iteratively updates the correction prompts. It summarizes the model's errors on the training set and uses these summaries to guide self-correction on the development set.
\end{itemize}

\section{Ablation Study for Skeleton-similarity Filtering}
\begin{table}[t]
\footnotesize
\resizebox{\columnwidth}{!}{%
\begin{tabular}{lcccccc}
\toprule
\textbf{Model} & & & \multicolumn{4}{c}{\textbf{All SQLs}}\\
\cmidrule(r){4-7} 
 & &  & \small \textbf{EX (\%)} &  &  & \small \textbf{VES (\%)} \\
\midrule

\multicolumn{1}{l}{\small {\textbf {GPT-3.5-turbo}}} & & & \small 41.98 &  &  & \small 47.78\\
\midrule

SQLCritic (w/o filtering)    & & & \small 43.47 & & & \small 48.32 \\
SQLCritic (w/ filtering)     & & & \small \textbf{48.94} &  &  & \small \textbf{51.32} \\

\midrule
\multicolumn{1}{l}{\small {\textbf {GPT-4o}}} & & & \small 49.87 &  &  & \small 52.43\\
\midrule

SQLCritic (w/o filtering)   & & & \small 50.79 & & & \small 53.43 \\
SQLCritic (w/ filtering)     & & & \small \textbf{54.58} &  &  & \small \textbf{55.98} \\

\bottomrule
\end{tabular}%
}
\caption{Performance of SQLCritic when trained with filtered data and without filtered data}
\label{tab:skeleton similarity}
\end{table}

This ablation study aims to evaluate the role of skeleton-similarity filtering in enhancing critique quality by analyzing its impact on model performance. Skeleton-similarity filtering ensures that training and critique generation focus on SQL queries with logical relevance, reducing noise caused by irrelevant or ill-constructed queries. To assess its effectiveness, we compared the performance of two different SQLCritic models trained with and without this filtering mechanism, using predictions from GPT-3.5-turbo and GPT-4o \cite{gpt4o}. As shown in Table \ref{tab:skeleton similarity}, removing skeleton-similarity filtering significantly degraded performance on both EX and VES metrics, indicating the importance of filtering for maintaining high-quality critiques that effectively guide the SQL refinement process.  

\section{Ablation Study for Critic DPO}
 We report the full results of our ablation study for Critic DPO in Tab. \ref{tab:full ablation}
\begin{table}[t]
\centering
\footnotesize
\resizebox{\columnwidth}{!}{%
\begin{tabular}{lcccccc}
\toprule
\textbf{Model} & & & \multicolumn{4}{c}{\textbf{All SQLs}}\\
\cmidrule(r){4-7} 
 & &  & \small \textbf{EX (\%)} &  &  & \small \textbf{VES (\%)} \\
\midrule
\multicolumn{1}{l}{\small {\textbf {GPT-3.5-turbo}}} & & & \small 41.98 &  &  & \small 47.78\\
\midrule

w/ SFT + Critic DPO    & & & \small \textbf{48.94} & & & \small \textbf{51.32} \\
w/ SFT + DPO     & & & \small 47.59 &  &  & \small 50.71 \\
w/ SFT     & & &  \small 45.94  &  &  & \small 48.88  \\

\midrule
\multicolumn{1}{l}{\small {\textbf {GPT-4o-mini}}} & & & \small 39.77 &  &  & \small 40.22\\
\midrule

w/ SFT + Critic DPO    & & & \small \textbf{47.79} & & & \small \textbf{49.12} \\
w/ SFT + DPO     & & & \small 46.87 &  &  & \small 48.01 \\
w/ SFT     & & &  \small 44.91  &  &  & \small 47.12  \\

\midrule
\multicolumn{1}{l}{\small {\textbf {o1-mini}}} & & & \small 47.85 &  &  & \small 49.66\\
\midrule

w/ SFT + Critic DPO    & & & \small \textbf{52.14} & & & \small \textbf{53.32} \\
w/ SFT + DPO     & & & \small 51.00 &  &  & \small 53.11 \\
w/ SFT     & & &  \small 49.29  &  &  & \small 52.33  \\

\midrule
\multicolumn{1}{l}{\small {\textbf {GPT-4o}}} & & & \small 49.87 &  &  & \small 52.43\\
\midrule

w/ SFT + Critic DPO    & & & \small \textbf{54.58} & & & \small \textbf{55.98} \\
w/ SFT + DPO     & & & \small 52.80 &  &  & \small 54.54 \\
w/ SFT     & & &  \small 51.92  &  &  & \small 53.34  \\

\midrule
\multicolumn{1}{l}{\small {\textbf {CodeS-7B}}} & & & \small 57.17 &  &  & \small 58.80\\
\midrule

w/ SFT + Critic DPO    & & & \small \textbf{59.34} & & & \small \textbf{64.28} \\
w/ SFT + DPO     & & & \small 59.19 &  &  & \small 63.26 \\
w/ SFT     & & &  \small 57.28  &  &  & \small 60.97  \\
\bottomrule
\end{tabular}%
}
\caption{Ablation study for the training method}
\label{tab:full ablation}
\end{table}





\end{appendices}
\end{document}